%% file: main.tex
\documentclass{article}
\usepackage{spconf,amsmath,graphicx,hyperref}
\usepackage{cleveref}
\usepackage{amssymb}
\usepackage{booktabs}
\usepackage{multirow}
\usepackage{siunitx}
\usepackage{graphicx}
\usepackage{tcolorbox}

\title{Entropy-Guided Data-Efficient Training for Multimodal \\ Reasoning Reward Models}
\makeatletter
\def\name#1{\gdef\@name{#1\\}}
\makeatother
\name{Shidong Yang$^{1,2}$\sthanks{Work done during the internship at AMAP, Alibaba Group.}, 
        Tongwen Huang$^{2}$\sthanks{Project leads and corresponding authors.},
        Hao Wen$^{2}$,
        Yong Wang$^{2}$\footnotemark[2],
        Li Chen$^{1}$,
        Xiangxiang Chu$^{2}$
        }
  
  \address{
  $^{1}$ School of Software, Tsinghua University \\
  $^{2}$~AMAP, Alibaba Group }

\begin{document}
\ninept
\maketitle
\begin{abstract}
\input{sections/abstract}
\end{abstract}

\input{sections/introduction}

\begin{figure*}[t!]
  \centering
  \includegraphics[width=1\linewidth]{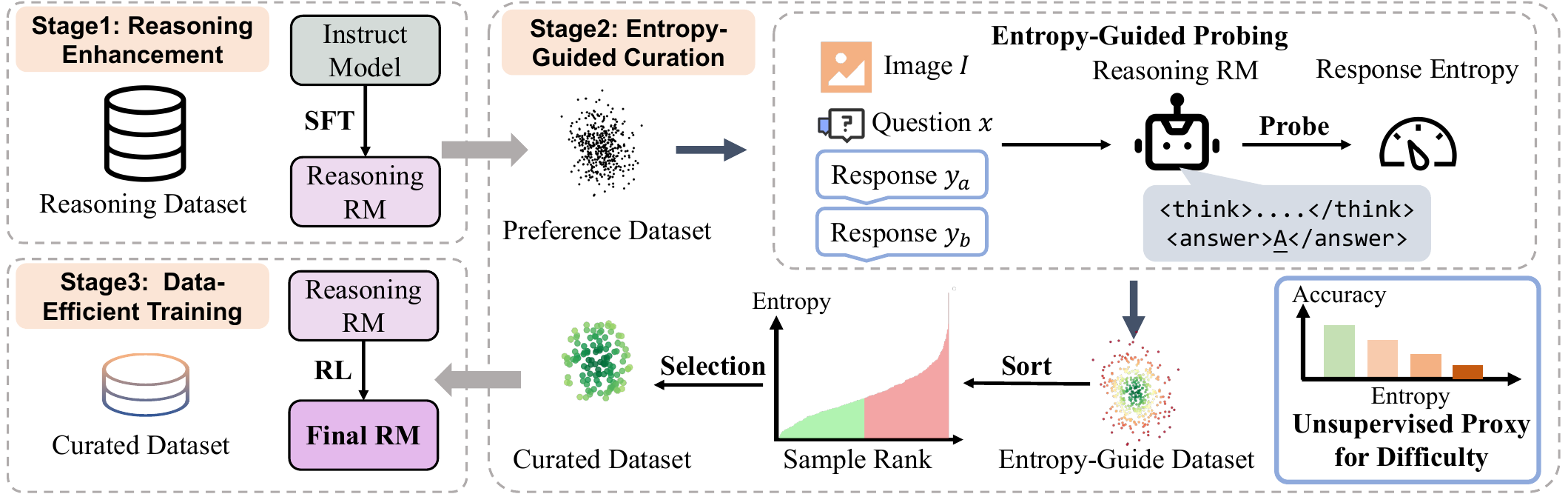}
  \caption{Overview of our proposed entropy-guided data-efficient training method.
  The process consists of three stages: (1) Reasoning Enhancement, where an instruction model is fine-tuned on high-quality reasoning trajectories; (2) Entropy-Guided Curation, where the reasoning-enhanced model prunes a preference dataset by identifying high-entropy samples. In this stage, the entropy probed by the reasoning reward model serves as a proxy for sample difficulty and noise; and (3) Data-Efficient Training, where the final model is trained on the curated dataset via reinforcement learning, following an easy-to-hard progression where samples are introduced in order of increasing entropy.
  } 
  \vspace{-0.25cm}
  \label{fig:train_process}
\end{figure*}
\input{sections/method}

\input{sections/experiment}

\input{sections/conclusion}

\bibliographystyle{IEEEbib}
\bibliography{refs}

\end{document}

%% file: sections/abstract.tex
Multimodal reward models are crucial for aligning multimodal large language models with human preferences. Recent works have incorporated reasoning capabilities into these models, achieving promising results. However, training these models suffers from two critical challenges: (1) the inherent noise in preference datasets, which degrades model performance, and (2) the inefficiency of conventional training methods, which ignore the differences in sample difficulty. In this paper, we identify a strong correlation between response entropy and accuracy, indicating that entropy can serve as a reliable and unsupervised proxy for annotation noise and sample difficulty. Based on this insight, we propose a novel \textbf{E}ntropy-\textbf{G}uided \textbf{T}raining (\textbf{EGT}) approach for multimodal reasoning reward models, which combines two strategies: (1) entropy-guided data curation to mitigate the impact of unreliable samples, and (2) an entropy-guided training strategy that progressively introduces more complex examples. Extensive experiments across three benchmarks show that the EGT-trained model consistently outperforms state-of-the-art multimodal reward models.

%% file: sections/introduction.tex
\section{Introduction}
\label{sec:intro}

Aligning Multimodal Large Language Models (MLLMs) with human preferences is a critical challenge~\cite{liu2024mmbench}. To address this, the Multimodal Reward Model (MRM) serves as a fundamental component~\cite{bai2022training}, which facilitates the high-quality data selection~\cite{rlaif_vs_rlhf} and provides reward signals for reinforcement learning~\cite{bai2022training}. Recent advancements in reasoning models~\cite{jaech2024openai, deepseekr1} have inspired the development of reward models referred to as Reasoning Reward Models~\cite {rm-r1, r1-reward}. These models incorporate reasoning trajectories to improve explanatory depth and utilize test-time scaling to boost performance.

\begin{figure}[t]
  \includegraphics[width=1.0\linewidth]{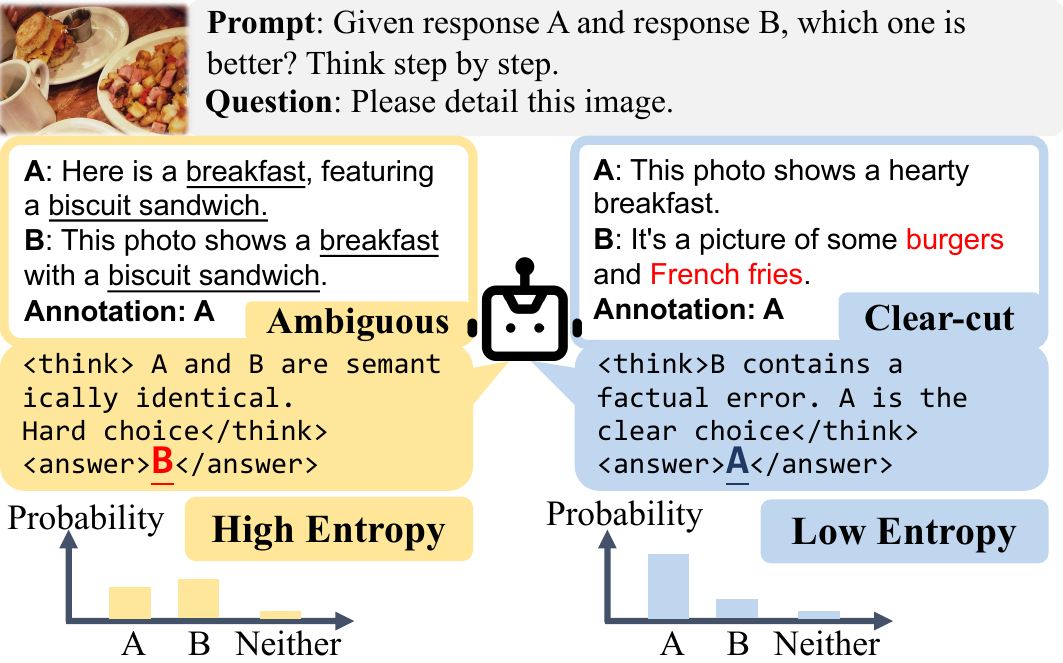}
  \caption{Response entropy can serve as a proxy for challenging and noisy samples. Left: An ambiguous sample results in a high-entropy output distribution. Right: A sample with a clear factual error allows for a confident, low-entropy decision.
  }
  \vspace{-0.25cm}
  \label{fig:case}
\end{figure}

However, the performance and efficiency of reasoning reward models are constrained by two fundamental issues: \textit{(1) data quality and robustness}~\cite{whatmakesgooddataforalign,secretsRM2}, and \textit{(2) training strategy and efficiency}~\cite{pattnaik2024enhancing,cao2025process,tian2025reinforcement,li2025adacurl}. Large-scale preference datasets often suffer from noise, such as ambiguous annotations where preferences are hard to discern. These unreliable data can impair the model's robustness during training and potentially cause performance degradation~\cite{gao2024impact, NEURIPS2020_1f89885d}. Second, conventional training methods adopt uniform random sampling of data, assuming all samples share equal importance and learning difficulty. This one-size-fits-all approach overlooks inherent variations in sample complexity, resulting in inefficient model training, particularly when handling challenging samples.

\begin{figure}[t]
  \centering
  \includegraphics[width=1.0\linewidth]{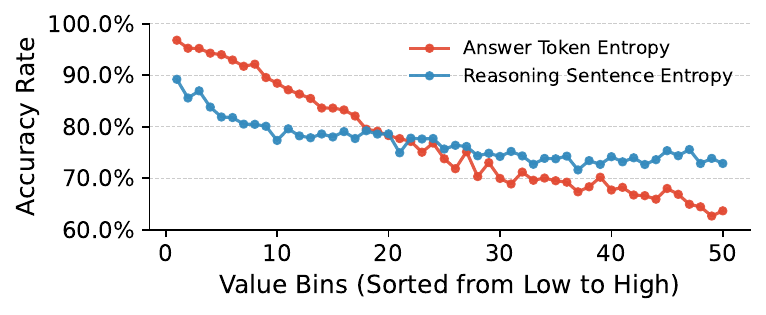}
  
  \caption{
  Correlation between response entropy and accuracy on a large-scale (80{,}000 samples) preference dataset. Samples are binned by their response entropy. The accuracy rate per bin reveals a clear downward trend: higher entropy correlates with lower accuracy.
  }
  \vspace{-0.25cm}
  \label{fig:ans_entropy}
\end{figure}

To address these challenges, we perform a rigorous analysis of the response probability distribution and identify response entropy as an effective indicator of both sample difficulty and noise. As illustrated in~\Cref{fig:case}, inherently ambiguous samples, which are particularly challenging for the model to evaluate, consistently exhibit high response entropy. We therefore hypothesize a negative correlation between response entropy and the accuracy of a reasoning reward model. To validate this, we construct a multimodal reasoning reward model and design two specific metrics derived from its structured output: \textit{reasoning sentence entropy} and \textit{answer token entropy}. Experimental validation conducted on a preference dataset supports our hypothesis (shown in~\Cref{fig:ans_entropy}), revealing an inverse relationship between entropy levels and model accuracy. Crucially, this proposed entropy-based method operates without the need for labeled data, offering a scalable and practical alternative compared to accuracy-based evaluations, which require ground-truth labels.

Leveraging the entropy-guided proxy, we propose \textbf{EGT}, an \textbf{E}ntropy-\textbf{G}uided data-efficient \textbf{T}raining approach for multimodal reasoning reward models. EGT integrates two strategies: (1) entropy-guided data curation, which constructs a compact, high-quality dataset by pruning high-entropy (ambiguous or extremely difficult) samples, and (2) a low-to-high entropy training strategy, which trains the model by progressively introducing samples of increasing complexity. 
Extensive experiments across three benchmarks demonstrate that our model consistently outperforms state-of-the-art models.

Our main contributions can be summarized as follows: (1) we demonstrate that the response entropy of a reasoning reward model serves as a reliable proxy for both sample difficulty and annotation noise in preference datasets; (2) we propose EGT, an entropy-guided data-efficient training framework that integrates entropy-driven data curation with a progressive training strategy to optimize the learning of multimodal reasoning reward models efficiently; (3) our EGT-trained reward models outperform previous approaches on three widely used multimodal reward benchmarks.

%% file: sections/method.tex
\section{Method}

\subsection{Task Definition}
Our framework is illustrated in~\Cref{fig:train_process}, which comprises three stages. Prior to delving into the specifics of the method, we formalize the task and define the key concepts essential to multimodal reward model training.

We define a reward model $\pi_\theta$, trained on a preference dataset $\mathcal{D} = \{(I_i,x_i, y_{a,i}, y_{b,i}, l_i)\}_{i=1}^N$. Each sample in $\mathcal{D}$ contains an image $I$, a query $x$, a pair of responses $(y_a, y_b)$, and a ground-truth label $l$. 
Given an input tuple $(I, x, y_a, y_b)$, the model generates an output $O$, from which the predicted label $\hat{l}$ is derived. The optimization objective of the training is formulated as follows:
\begin{equation}
    \max_{\pi_\theta}\;
\mathbb{E}_{(I, x,y_a,y_b,l)\sim\mathcal{D},\;
          \hat{l}\sim \pi_\theta\!\bigl(O \mid I,x,y_a,y_b\bigr)}
\Bigl[\,\mathbb{I}\!\bigl(\hat{l}=l\bigr)\Bigr],
\end{equation}
where $\mathbb{I}(.)$ is an indicator function. This objective aligns the model with human preferences by rewarding correct predictions.

\subsection{Enhancing RM with Reasoning Capabilities}
To enhance the reasoning capability of a base instruction model, we leverage an advanced reasoning model (e.g., Gemini 2.5 Pro) to generate detailed reasoning trajectories $r_i$ for each preference sample $d_i = (I_i, x_i, y_{a,i}, y_{b,i})$ in $\mathcal{D}$. Generation follows a strict fidelity filter: the model may attempt up to three times without access to the ground truth. If all attempts fail, the sample is discarded; otherwise, we retain the first successful trajectory. The retained pairs form a high-quality reasoning set $\mathcal{D}_{\text{sft}}$, on which we fine-tune the instruction model by minimizing the negative log-likelihood objective:
\begin{equation}
\mathcal{L}_{\text{refined}}(\theta) = -\mathbb{E}_{(d_i, y_i) \sim \mathcal{D}_{\text{sft}}} \left[ \log p_{\theta}(r_i,l_i \mid d_i) \right].
\end{equation}

\subsection{Entropy-Guided Data Curation}

Preference datasets inevitably contain noise. To address this, we introduce an entropy-guided curation framework that adopts response entropy to prune unreliable and extremely difficult samples to form a curated set $\mathcal{D}_{\text{curated}}$.

The framework begins with entropy-guided probing, a process where the reasoning reward model $\pi_\theta$ generates an output sequence $O_i = (t_1, t_2, \dots, t_L)$ for each input $(I_i, x_i, y_{a,i}, y_{b,i},l_i)$ to compute its response entropy, where $t_j$ denotes the $j$-th token in the generated response.
Given the structured nature of this output, which contains both reasoning steps and a final answer (as shown in \Cref{fig:case}), we decompose the total response entropy into two key components:

\noindent\textbf{Answer Token Entropy.} 
    In reasoning reward models, the final answer corresponds to a single token within the vocabulary $V$. To compute the entropy, we first extract the model's output logits $\mathbf{p}_i$ and apply the softmax function, yielding a probability distribution over the entire vocabulary. For a generated sequence $O_i$, the entropy corresponding to the answer token is defined as:
    \begin{equation}
e_i^{\text{answer}}
= - \sum\nolimits_{k=1}^{|V|} p_{i,k}\,\log p_{i,k},
    \end{equation}
    where $|V|$ represents the size of the vocabulary, and $p_{i,k}$ is the probability of the $k$-th vocabulary token at the answer position.
    
\noindent\textbf{Reasoning Sentence Entropy.} To quantify the uncertainty in the reasoning process, we compute the average token entropy across the entire sequence as:
    \begin{equation}
    e_i^{\text{reasoning}} = \frac{1}{L} \sum\nolimits_{j=1}^{L} \Bigl( - \sum\nolimits_{k=1}^{|V|} p_{i,j,k}\, \log p_{i,j,k} \Bigr),
    \end{equation}
    where $L$ denotes the length of the sequence, and $p_{i,j,k}$ is the probability of the $k$-th vocabulary token at position $j$, conditioned on the preceding sequence $(t_1,\dots,t_{j-1})$. 

By combining these two metrics, we formulate a composite entropy score $e_i = f( e_i^{\text{answer}}, e_i^{\text{reasoning}})$, where $f$ is a function that balances the two components. The design of $f$ is detailed in~\Cref{sec:exp_analysis}. 

This entropy score provides a direct criterion for data curation. To prune excessively difficult or potentially noisy data, we construct the curated training set as $\mathcal{D}_{\text{curated}} = \{ d_i \in D \mid e_i < q_p \}$, where $q_p$ represents the $p$-th percentile of the entropy distribution. 

\subsection{Data-Efficient Training}
\label{sec:rl_training}
After curating the high-quality dataset $\mathcal{D}_{\text{curated}}$, the model is further optimized through reinforcement learning, which incorporates our entropy-based ranking strategy and a composite reward function.

\noindent\textbf{Entropy-based Ranking.} 
The entropy scores calculated during the probing stage provide a natural, unsupervised proxy for sample difficulty.
We leverage this insight by implementing an entropy-based training curriculum rather than uniform sampling. Specifically, the training is sequenced from low-entropy samples, representing clear-cut cases, to high-entropy ones, which are more complex. This training strategy enables the model to establish a robust foundation on simpler data before tackling more complex examples, resulting in more stable and efficient optimization.

\noindent\textbf{Reward Design.} Following previous work~\cite{deepseekr1}, we utilize a rule-based reward function $R$ that integrates accuracy, format, and logic terms, defined as $R = R_{\text{acc}} (1 + \alpha  R_{\text{logic}}) + \beta  R_{\text{format}}$. The accuracy reward $R_{\text{acc}}$ is one if the answer is correct and zero otherwise. This base score is subsequently modulated by $R_{\text{logic}}$, a term that assigns 1 for reasoning that logically supports a correct answer and imposes a penalty of -1 for a misaligned trajectory. Additionally, an independent format reward $R_{\text{format}}$ is granted a value of 1 if the output strictly adheres to the prescribed structure. The balancing hyperparameters $\alpha$ and $\beta$ are both set to 0.5 in our experiments.

%% file: sections/experiment.tex
\section{Experiments}
\input{tables/tab_merge}
\subsection{Experiments Setup}
\noindent\textbf{Dataset.} 
For the SFT stage, we generate 100,000 reasoning trajectories from five publicly multimodal preference datasets~\cite{ mm-rlhf, wildvision, povid, rlhf-v, li-etal-2024-vlfeedback} and use them to instruction-tune the base model, enhancing its ability to reward modeling and improve its reasoning capabilities. For the RL stage, following previous work~\cite{r1-reward}, we adopt a challenging dataset of 17,000 preference pairs. This dataset contains samples that require multiple attempts, even for advanced models like GPT-4o to solve correctly, implying a mixture of complex cases and noisy artifacts, making it an ideal candidate for refinement using our proposed entropy-guided data curation.

\noindent\textbf{Implementation Details.}
All experiments are conducted on the Qwen2.5-VL-7B-Instruct~\cite{bai2025qwen2}, using 8$\times$H20 GPUs. For the SFT stage, we utilize LlamaFactory~\cite{zheng2024llamafactory} and fine-tune for one epoch with a batch size of 256 and a learning rate of 1e-5. For the subsequent RL phase, we employ our entropy-based method, selecting the 2,500 lowest-entropy samples (based on the answer token) from the whole RL dataset to form a curated training set. The model is then trained with StableReinforce for $20$ epochs within the OpenRLHF~\cite{hu2024openrlhf}. The training batch size is $224$ and the learning rate is 1e-6. At each epoch, the training data are sorted in ascending order of entropy.

\input{tables/abl_filter_l2h}

\vspace{-0.2cm}
\subsection{Main Results}
We conduct experiments on three widely used multimodal reward modeling benchmarks~\cite{mm-rlhf, vlrewardbench, multimodalrewardbench} and achieve competitive performance across both open-/closed-source models and specialized reward models as detailed in~\Cref{tab:merge_with_avg}. To further validate the effectiveness of each component in our approach, we conduct ablation studies on VL-RewardBench, as shown in~\Cref{tab:abl_filter_l2h}. Based on the experimental results, we highlight our main findings in the following:

\noindent\textbf{State-of-the-Art Performance.} Our method establishes a new state-of-the-art in multimodal reward modeling, outperforming the previous leading approach, R1-Reward, by 3.88\% on average. Remarkably, these performance gains are consistent across all evaluated datasets (4.26\% on VL-RewardBench, 2.10\% on Multimodal Reward Bench, and 3.53\% on MM-RLHF Reward Bench), demonstrating the robust generalization capability of our model.

\noindent\textbf{Data-efficient Training.} Our entropy-guided data curation method successfully identifies a reliable subset of the data. Notably, we retain only 2,500 samples for training. Even with this drastic reduction in data size, our trained model achieves performance comparable to training with the full dataset. This reduction in sample requirement lowers computational costs, making large-scale reward model training more accessible and sustainable. 

\noindent\textbf{Effectiveness of Entropy as a Difficulty Proxy.} The performance of the entropy-guided set over a full dataset proves that entropy can serve as a reliable proxy for data difficulty and noise. The act of removing high-entropy samples aims to remove ambiguous or extremely challenging samples that can confuse the model. 
Entropy-guided pruning thus enables the construction of cleaner, more coherent training subsets that facilitate robust and generalizable learning. The demonstrated effectiveness of entropy-guided pruning also opens up new possibilities for adopting advanced training strategies like adaptive sampling in future research.

\vspace{-0.2cm}
\subsection{Analysis}
\label{sec:exp_analysis}
In this section, we conduct a series of empirical analyses to evaluate our proposed EGT framework fully. Our investigation involves benchmarking EGT on VL-RewardBench against alternative data selection methods, analyzing the design of various entropy calculation strategies, assessing the impact of training data scale, and the contributions of data from different entropy levels.

\input{tables/abl_strategy}

\input{tables/abl_entropy}

\input{tables/abl_entropy_level}

\noindent\textbf{Comparison with Alternative Selection Strategies.}
The results in \Cref{tab:abl_strategy} show that our entropy-based selection consistently outperforms both random and accuracy-based baselines. Compared to random selection, our method is more efficient in identifying valuable samples. Furthermore,  unlike supervised accuracy-based selection, our approach is entirely unsupervised, making it more flexible.

\noindent\textbf{Evaluation of Entropy Score Design.}
We explore various implementations of entropy strategies in~\Cref{tab:abl_entropy}, revealing that computing the entropy score based on the answer token yields the best results. In contrast, calculating the entropy over entire reasoning sentences provides minimal information gain. This underperformance may be due to the longer sentence lengths, where averaging dilutes the information signal, making it less robust compared to the more focused probability distribution associated with individual answer tokens.

\begin{figure}
  \centering
  \includegraphics[width=1.0\linewidth]{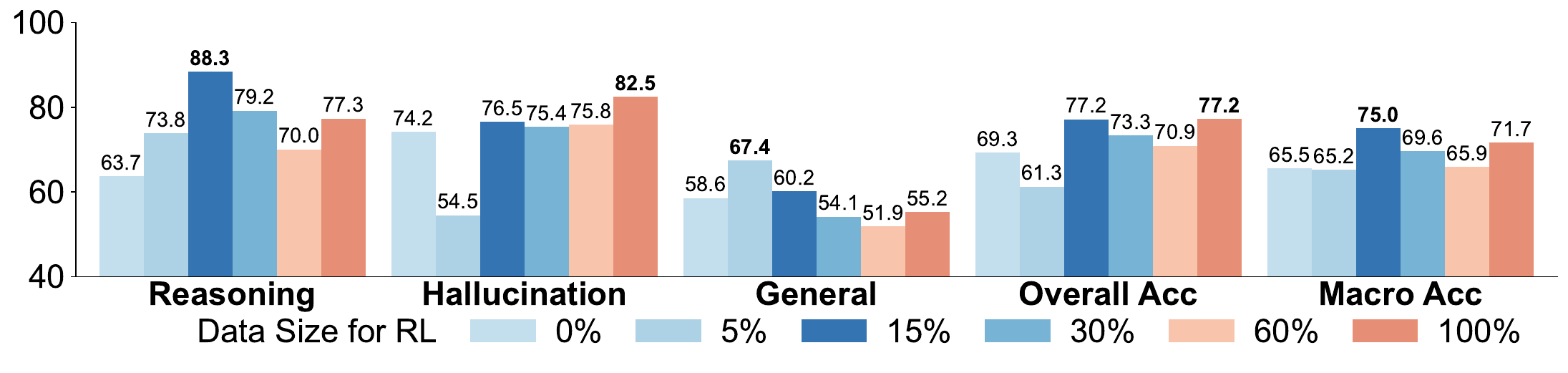}
  \caption{Performance comparison with different data scales. 
  }
    \vspace{-0.6cm} 
  \label{fig:data_scale}
\end{figure}

\noindent\textbf{Impact of Data Size on Performance.}
We analyze the impact of data quantity by training on subsets of the lowest-entropy data, ranging from 0\% (the SFT baseline) to 100\%. These subsets are selected in ascending order by answer token entropy. As shown in~\Cref{fig:data_scale}, using the 15\% lowest-entropy data achieves performance comparable to using the entire dataset. This result further confirms the presence of redundancy and noise in the training dataset, as well as the effectiveness of our proposed method.

\noindent\textbf{The Performance of Different Entropy Levels.}
To further validate our hypothesis that high-entropy data can introduce noisy and extremely difficult samples for training, we perform an ablation study on data at different entropy levels. This experiment investigates whether the type of data, as defined by its entropy level, is the critical factor. 
We partition the dataset into three 2,500-sample subsets based on answer token entropy: Low, Mid (around the median), and High.
As presented in~\Cref{tab:abl_entropy_levels}, the model trained on the Low-Entropy subset outperforms the others. In contrast, training on the high-entropy subset leads to poor performance, suggesting that such data may introduce noise or conflicting signals that impede learning.

%% file: tables/tab_merge.tex
\begin{table*}[htbp]
\centering
\small
\caption{Results on three multimodal reward benchmarks: VL-RewardBench (VL-Reward), Multimodal RewardBench (Multimodal), and MM-RLHF-RewardBench (MM-RLHF). 
\textbf{Bold} indicates the best, with the superscript indicating the improvement over the \underline{second-best} result (underlined). For clarity, we report the overall accuracy for each benchmark. The \textbf{Avg. Gain} is relative to the GPT-4o baseline. }
\begin{tabular}{l|c|ccc|c|c}
\toprule
\textbf{Model} & \textbf{\# Param} & \textbf{VL-Reward} & \textbf{Multimodal} & \textbf{MM-RLHF} & \textbf{Avg.} & \textbf{Avg. Gain} \\
\midrule
\multicolumn{7}{c}{\textbf{\textit{Proprietary Models}}} \\ 
\midrule
GPT-4o (2024-08-06) & -- & 65.80 & 70.80 & 58.23 & 64.94 & -- \\
Claude-3.7-Sonnet (2025-02-24) & -- & 66.31 & 71.90 & \underline{82.35} & 73.52 & $\uparrow$8.58 \\
\midrule
\multicolumn{7}{c}{\textbf{\textit{Open-source Models}}} \\ 
\midrule
SliME~\cite{zhang2024benchmarking} & 7B & 19.04 & 42.00 & 17.10 & 26.05 & $\downarrow$38.89 \\
VITA-1.5~\cite{fu2025vita} & 7B & 16.48 & 53.60 & 20.58 & 30.22 & $\downarrow$34.72 \\
Qwen2-VL-72B~\cite{bai2025qwen2} & 72B & 39.50 & 70.90 & 48.23 & 52.88 & $\downarrow$12.06 \\
\midrule
\multicolumn{7}{c}{\textbf{\textit{Specialized Reward Models}}} \\ 
\midrule
MM-RLHF-Reward~\cite{mm-rlhf} & 7B & 50.15 & 67.10 & 82.00 & 66.42 & $\uparrow$1.48 \\
IXC-2.5-Reward~\cite{zang2025internlm} & 7B & 65.80 & 66.60 & 71.18 & 67.86 & $\uparrow$2.92 \\
R1-Reward~\cite{r1-reward} & 7B & \underline{72.89} & \underline{82.20} & 80.59 & \underline{78.56} & $\uparrow$\underline{13.62} \\
EGT (Ours) & 7B & \textbf{77.15} & \textbf{84.30} & \textbf{85.88} & \textbf{82.44} & $\uparrow$\textbf{17.50} \\
\bottomrule
\end{tabular}
\label{tab:merge_with_avg}
\end{table*}

%% file: tables/abl_filter_l2h.tex
\begin{table}
\centering
\small
\caption{Ablation study of different training strategies on VL-RewardBench. 
``Full RL'' is the baseline using the entire dataset. ``+ Selection'' applies RL on the 2500 lowest-entropy samples. ``+ Selection + Sort'' further refines this process by arranging the selected samples in ascending order of entropy.}
\begin{tabular}{@{}lcccc@{}}
\toprule
\small
\textbf{Method} & \textbf{Reasoning} & \textbf{Hallu.} & \textbf{General} & \textbf{Overall} \\ \midrule
SFT & 63.72 & 74.23 & 58.56 & 69.29 \\
+ Full RL & 68.77 & 76.50 & 54.70 & 71.37 \\
+ Selection & 84.54 & 68.49 & 58.56 & 71.13 \\
+ Selection \& Sort & \textbf{88.33} & \textbf{76.50} & \textbf{60.20} & \textbf{77.15} \\ \bottomrule
\end{tabular}
\label{tab:abl_filter_l2h}
\end{table}

%% file: tables/abl_strategy.tex
\begin{table}
\centering
\small
\caption{Ablation study of data selection strategies. Accuracy represents selecting samples with the lowest correctness scores, while the random strategy employs random sampling from the dataset.}
\begin{tabular}{@{}lcccc@{}}
\toprule
\textbf{Method} & \textbf{Reasoning} & \textbf{Hallu.} & \textbf{General} & \textbf{Overall} \\ \midrule
Random & 66.88 & 75.03 & 53.04 & 69.77 \\
Accuracy & 67.19 & 73.16 & 53.59 & 68.61 \\
Entropy & \textbf{88.33} & \textbf{76.50} & \textbf{60.20} & \textbf{77.15} \\ \bottomrule
\end{tabular}
\label{tab:abl_strategy}
\end{table}

%% file: tables/abl_entropy.tex
\begin{table}[htbp]
\centering
\small
\caption{Impact of entropy-based selection criteria. The Mix uses the product of the sentence entropy and answer entropy of each sample as an indicator. In all cases, the 2,500 samples with the lowest entropy indicator are selected. }
\begin{tabular}{@{}lcccc@{}}
\toprule
\textbf{Method} & \textbf{Reasoning} & \textbf{Hallu.} & \textbf{General} & \textbf{Overall} \\ \midrule
Sentence & 73.19 & 72.76 & 48.62 & 69.37 \\
Answer & \textbf{88.33} & \textbf{76.50} & \textbf{60.20} & \textbf{77.15} \\ 
Mix & 84.23 & 74.23 & 57.46 & 74.34 \\
\bottomrule
\end{tabular}
\label{tab:abl_entropy}
\end{table}

%% file: tables/abl_entropy_level.tex
\begin{table}
\centering
\small
\caption{Performance under different entropy levels.}
\begin{tabular}{@{}lcccc@{}}
\toprule
\textbf{Method} & \textbf{Reasoning} & \textbf{Hallu.} & \textbf{General} & \textbf{Overall} \\ \midrule
Low-Entropy    & \textbf{88.33} & \textbf{76.50} & \textbf{60.20} & \textbf{77.15} \\ 
Mid-Entropy & 65.30  & 73.43 & 50.28 & 68.00   \\
High-Entropy   & 65.30  & 73.56 & 56.35 & 68.97 \\  \bottomrule
\end{tabular}
\label{tab:abl_entropy_levels}
\end{table}

%% file: sections/conclusion.tex
\section{Conclusion}
We introduce \textbf{EGT}, an \textbf{E}ntropy-\textbf{G}uided data-efficient \textbf{T}raining framework for multimodal reasoning reward models.
Our approach is built on a key insight: a strong correlation exists between response entropy and accuracy. This indicates that the response entropy serves as a reliable, unsupervised proxy for both sample difficulty and annotation noise. EGT leverages this principle through a combination of entropy-guided data curation and a low-to-high entropy curriculum, enabling more efficient and robust model training. We apply EGT to a multimodal reasoning reward model, and extensive experiments show that our approach achieves competitive performance.